# OPFData: Large-scale datasets for AC optimal power flow with topological perturbations

Sean Lovett[1,*], Miha Zgubič[1,*], Sofia Liguori[1], Sephora Madjiheurem[1], Hamish Tomlinson[1], Sophie Elster[1], Chris Apps[1], Sims Witherspoon[1] and Luis Piloto[1]
[1]Google DeepMind, *Equal contribution

Solving the AC optimal power flow problem (AC-OPF) is critical to the efficient and safe planning and operation of power grids. Small efficiency improvements in this domain have the potential to lead to billions of dollars of cost savings, and significant reductions in emissions from fossil fuel generators. Recent work on data-driven solution methods for AC-OPF shows the potential for large speed improvements compared to traditional solvers; however, no large-scale open datasets for this problem exist. We present the largest readily-available collection of solved AC-OPF problems to date. This collection is orders of magnitude larger than existing readily-available datasets, allowing training of high-capacity data-driven models. Uniquely, it includes topological perturbations - a critical requirement for usage in realistic power grid operations. We hope this resource will spur the community to scale research to larger grid sizes with variable topology.

> The **OPFData** datasets are distributed as JSON files in a public Google Cloud bucket `gridopt-dataset` along with the LICENSE and README files. We also provide loading utilities for torch geometric.

## 1. Introduction

Power grids are among the largest, most complex, and most critical pieces of infrastructure. They are also major contributors to greenhouse gas emissions, with the global power sector accounting for almost 40% of all energy-related $CO_2$ emissions [1]. Operating power grids efficiently and securely requires solving variants of the alternating current optimal power flow (AC-OPF) problem to determine optimal unit commitment, power dispatch, and related quantities, subject to physical constraints, and often subject to preventative security constraints [2]. In its general form this is a non-linear, non-convex, mixed-integer constrained optimisation problem. Typical methods for solving such problems are too computationally expensive or lack sufficient robustness for real-time application on large grids, necessitating the use of approximate formulations such as the linearised DC-OPF [3, 4]. Approximate formulations are robust and fast to solve but can introduce substantial inefficiencies, both economic [5] and emissions-related [6], and typically violate the true constraints of the power network [7], requiring further post-processing which can degrade optimality of the solution. Higher penetration of renewable generation is expected to increase the difficulty of these problems, as it leads to lower-inertia grids and a higher reliance on uncertain intermittent generation [8, 9, 10], as well as increased grid congestion requiring balancing actions [11].

There is therefore the potential for substantial economic, security, and emissions benefits from solving true AC-OPF formulations rather than approximations. This is also a particularly high-leverage area for applying machine learning to tackle climate change [12]. Machine learning (ML)-based approaches for solving OPF have attracted interest due to their potential for important speed improvements on a variety of OPF-related problems; a recent review can be found in Khaloie et al. [13]. However, for ML-based OPF solutions to be applied in real power grid operations, they must be reliable and robust. A common shortcoming in this regard is infeasibility with respect to constraints of the original problem formulation. This can be tackled in a number of ways, for example by post-processing model outputs [14] or directly during learning [15]. Another basic requirement for operations is robustness to grid topology variation, either for contingency analysis [16], due to planned addition or removal of grid components, network topology switching, or unplanned outages [17]. Graph-based models are naturally suited to this requirement, since their structure reflects the relational structure of the data [18, 19, 20].

Despite the interest in the field, there are few standardised datasets available. As a result, the typical approach is for researchers to generate their own; due to computational complexity this can involve substantial investments of compute time for large grids or where discrete decision variables are present. One exception known to the authors is the freely-available OPF-Learn dataset [21], which is generated using a method designed to maximise the variety of active constraint sets in the solutions; pre-generated datasets are available online for small grid sizes (up to 118 buses). OPF-Learn and other recent research into efficient sampling of the feasible set [e.g. 22, 23] shows promise for improving model results for a given dataset size. However, it is an open question how best to extend these techniques to more general settings such as datasets with topological variation, where each example in the dataset could have a different feasible set. A second freely-available dataset is the TAS-97 dataset, which contains realistic load patterns on a realistic grid modelling Tasmania's electricity network [24]. This dataset captures realistic correlations between loads; however, the number of samples (7284) and grid size (97 buses) are both small. Our findings, and those of the review by Khaloie et al. [13], are that adaptability to network variation, scalability to large grids and large datasets, and the lack of standardized datasets and benchmarking platforms remain key challenges in the field.

## 2. The OPFData dataset

In this paper we introduce **OPFData**, a collection of open datasets for research into ML methods for solving AC-OPF, aiming to assist with addressing some of the challenges identified above. Each example within the datasets is a self-contained OPF problem with solution, allowing for complete flexibility to represent different grid structures and properties between examples. It is our view that models for operational application must be able to gracefully handle and learn from variable grid topology specifications. This is true from both from a pragmatic standpoint—because the grid constantly varies—and an application-specific one, for example for contingency screening. Hence we include datasets with topological perturbations.

**OPFData** consists of 300k solved AC-OPF problems for each of a range of grids, making it the largest of such datasets openly available. Grid sizes range up to 13659 buses. Problems are based on the base test cases from the widely-used PGLib-OPF library [31]. We select a number of commonly-used grids spanning a range of sizes, and for each grid we present two datasets:

- `FullTop`: This is a simple model for a fixed grid with variable load conditions. We multiply each active and reactive load value independently by a random number drawn uniformly from $[0.8, 1.2]$, similar to e.g. Fioretto, Mak, and Van Hentenryck [15].
- `N-1`: This is a simple model for a variable grid with variable load conditions. We perturb load as above, and additionally (with probability 0.5) choose a single generator uniformly at random to drop, or (with probability 0.5) choose a single line/transformer uniformly at random to drop. Components which are dropped are removed entirely from the network specification. We do not drop generators connected to reference buses, and we do not drop a component if doing so would result in a disconnected graph.

Some of the perturbations above lead to infeasible problems, which are discarded. Out of the five core variability factors identified in Popli et al. [17] for building ML-OPF datasets (load distribution, load power factor, generator outages, line outages, generator costs), the above covers all but generator costs, which we leave to future iterations. The datasets are summarised in table 1; a detailed description is given in appendix A.

The **OPFData** datasets are distributed as JSON

Table 1 | Summary of the **OPFData** datasets, detailing in each case the base scenario from PGLib-OPF with original source citation, and number of buses $|\mathcal{N}|$, generators $|\mathcal{G}|$, loads $|\mathcal{D}|$, shunts $|\mathcal{S}|$, and edges $|\mathcal{E}_l|$ (AC lines), $|\mathcal{E}_t|$ (transformers). Each dataset consists of 300k examples and has `FullTop` and `N-1` variants (so there are 20 datasets in total).

| Scenario | $|\mathcal{N}|$ | $|\mathcal{G}|$ | $|\mathcal{D}|$ | $|\mathcal{S}|$ | $|\mathcal{E}_l|$ | $|\mathcal{E}_t|$ |
| --- | --- | --- | --- | --- | --- | --- |
| pglib_opf_case14_ieee [25] | 14 | 5 | 11 | 1 | 17 | 3 |
| pglib_opf_case30_ieee [25] | 30 | 6 | 21 | 2 | 34 | 7 |
| pglib_opf_case57_ieee [25] | 57 | 7 | 42 | 3 | 63 | 17 |
| pglib_opf_case118_ieee [25] | 118 | 54 | 99 | 14 | 175 | 11 |
| pglib_opf_case500_goc [26, 27] | 500 | 171 | 281 | 31 | 536 | 192 |
| pglib_opf_case2000_goc [26, 27] | 2000 | 238 | 1010 | 124 | 2737 | 896 |
| pglib_opf_case4661_sdet [28] | 4661 | 724 | 2683 | 696 | 4668 | 1329 |
| pglib_opf_case6470_rte [29] | 6470 | 761 | 3670 | 73 | 7426 | 1579 |
| pglib_opf_case10000_goc [26, 27] | 10000 | 2016 | 3984 | 510 | 10819 | 2374 |
| pglib_opf_case13659_pegase [29, 30] | 13659 | 4092 | 5544 | 8754 | 13792 | 6675 |

files, available in a public Google Cloud bucket (`gs://gridopt-dataset/`), and are agnostic to ML frameworks or model architectures. For convenience we additionally supply utilities for working with **OPFData** in PyTorch Geometric [32], a popular ML framework for graph learning. Listing 1 shows an example of training a simple graph neural network on a batched **OPFData** dataset in a few lines of code; for more details see the PyG documentation. In Piloto et al. [33] we apply a graph-based model written in JAX [34] and jraph [35] to equivalent datasets, including detailed consideration of constraint satisfaction metrics[1].

## 3. Summary and future work

We have presented **OPFData**, an openly-available collection of datasets of solved AC-OPF problems in a format amenable to common ML workflows. Taken together, **OPFData** is the largest of such solved and openly-available datasets in terms of number of examples, number of grids, and grid sizes. We hope to encourage researchers to explore this fascinating and important field, and particularly to scale methods to large grids with topological variations.

While we believe these datasets are useful as-is, there are several dimensions of improvement possible:

- A more efficient or representative exploration of the feasible space. One example could be a more sophisticated distribution of load perturbations, for example a truncated normal [14, 36]. Another could be a scheme such as the one presented by Joswig-Jones, Baker, and Zamzam [21], aiming to explore the space of active constraint sets more thoroughly than a simple perturbation.
- Further non-topological perturbations; for example perturbing generator capacities or line properties, or generator costs to generalise to different fuel prices.
- Further topological perturbations; for example dropping more than one component, adding additional lines, or re-configuring existing lines.
- Further output features. For example active constraint sets of the OPF solution could be used by classification models [13], or dual solutions could be used to learn models which output locational marginal prices [19].

## References

[1] IEA. *World Energy Outlook 2023*. Paris: IEA, 2023.

---

[1]Note that the `TopDrop` dataset from Piloto et al. [33] can be imitated with a 50-50 mix of the **OPFData** `FullTop` and `N-1` datasets.

```
!pip install pyg-nightly # Only necessary until PyG 2.6.0 is released.

import torch
import torch.nn.functional as F
from torch_geometric.nn import GraphConv, to_hetero
from torch_geometric.datasets import OPFDataset
from torch_geometric.loader import DataLoader

# Load the 14-bus OPFData FullTopology dataset training split and store it in the
# directory 'data'. Each record is a `torch_geometric.data.HeteroData`.
train_ds = OPFDataset('data', case_name='pglib_opf_case14_ieee', split='train')
# Batch and shuffle.
training_loader = DataLoader(train_ds, batch_size=4, shuffle=True)

# A simple model to predict the generator active and reactive power outputs.
class Model(torch.nn.Module):
  def __init__(self):
    super().__init__()
    self.conv1 = GraphConv(-1, 16)
    self.conv2 = GraphConv(16, 2)

  def forward(self, x, edge_index):
    x = self.conv1(x, edge_index).relu()
    x = self.conv2(x, edge_index)
    return x

# Initialise the model.
# data.metadata() here refers to the PyG graph metadata, not the OPFData metadata.
data = train_ds[0]
model = to_hetero(Model(), data.metadata())

with torch.no_grad(): # Initialize lazy modules.
  out = model(data.x_dict, data.edge_index_dict)
  # Train with MSE loss for one epoch.
  # In reality we would need to account for AC-OPF constraints.
  optimizer = torch.optim.Adam(model.parameters())
  model.train()

for data in training_loader:
  optimizer.zero_grad()
  out = model(data.x_dict, data.edge_index_dict)
  loss = F.mse_loss(out['generator'], data['generator'].y)
  loss.backward()
  optimizer.step()
```

Listing 1 | Example usage of **OPFData** with PyTorch Geometric.

## A. Detailed description of the datasets

**OPFData** is available in a public Google Cloud bucket (`gs://gridopt-dataset/`). For licensing and other general information see the LICENSE and README files in that bucket.

**OPFData** is based on the widely-used PGLib-OPF datasets [31]. AC-OPF problems were solved using the Julia language [37] with PowerModels.jl [38] and the Ipopt [39] and MUMPS [40] solvers. Figure 1 shows the AC-OPF formulation, taken from PowerModels.jl; see the `ACPPowerModel` in PowerModels.jl documentation for full details.

Each example in the dataset is a JSON file having the structure shown in listing 2. At the root are `grid`, representing the grid state (OPF inputs); `solution`, representing the OPF solution, and `metadata`, containing metadata about the example. The canonical dataset split for **OPFData** is a train / validate / test split of 0.9 / 0.05 / 0.05, resulting in 270k / 15k / 15k examples. The examples are i.i.d. and are taken sequentially, i.e. the train set consists of `example_{0-26999}.json`, the validation set of `example_{270000-284999}.json`, and the test set of `example_{285000-299999}.json`.

The `grid` and `solution` are split into nodes and edges. Nodes are split into several types containing feature matrices where the rows correspond to the entity number, and the columns to features. Edges are split into AC lines and transformers, and the `senders` (from) and `receivers` (to) additionally contain integer indices corresponding to the connected entities[2]:

- `ac_line` indicates a line connecting a bus to a bus;
- `transformer` indicates a transformer connecting a bus to a bus;
- `generator_link` indicates a line connecting a generator (sender) to a bus (receiver);
- `load_link` indicates a line connecting a load (sender) to a bus (receiver);

---

[2]The `generator`, `load` and `shunt` nodes are the "subnodes" from Piloto et al. [33], and `generator_link`, `load_link`, `shunt_link` the corresponding artificial (subnode-bus) edges.

```
{
    "grid": {
        "nodes": {
            "bus": ...,
            "generator": ...,
            "load": ...,
            "shunt": ...
        },
        "edges": {
            "ac_line": {
                "senders": ...,
                "receivers": ...,
                "features": ...
            },
            "transformer": {
                "senders": ...,
                "receivers": ...,
                "features": ...
            },
            "generator_link": {
                "senders": ...,
                "receivers": ...
            },
            "load_link": {
                "senders": ...,
                "receivers": ...
            },
            "shunt_link": {
                "senders": ...,
                "receivers": ...
            }
        },
        "context": ...
    },
    "solution": {
        "nodes": {
            "bus": ...,
            "generator": ...
        },
        "edges": {
            "ac_line": {
                "senders": ...,
                "receivers": ...,
                "features": ...
            },
            "transformer": {
                "senders": ...,
                "receivers": ...,
                "features": ...
                ]
            }
        }
    },
    "metadata": {
        "objective": ...
    }
}
```

Listing 2 | Structure of the dataset JSON files.

- shunt_link indicates a line connecting a shunt (sender) to a bus (receiver).

The features, in column order in each case, are given below. Symbols refer to the formulation in figure 1. These correspond to a subset of features in a PowerModels.jl network [38]. Unless otherwise specified, parameters are in per-unit and angles are in radians.

- grid.nodes.bus:

base_kv: Base voltage (kV).
bus_type: PQ (1), PV (2), reference (3), inactive (4).
vmin ($v_i^l$): Minimum voltage magnitude.
vmax ($v_i^u$): Maximum voltage magnitude.

- grid.nodes.generator:

mbase: Total MVA base.
pg: Initial real power generation as given in the pglib case.
pmin: ($\Re(S_k^{gl})$) Minimum real power generation.
pmax: ($\Re(S_k^{gu})$) Maximum real power generation.
qg: Initial reactive power generation as given in the pglib case.
qmin: ($\Im(S_k^{gl})$) Minimum reactive power generation.
qmax: ($\Im(S_k^{gu})$) Maximum reactive power generation.
vg: Initial voltage magnitude as given in the pglib case.
cost_squared: ($c_{2k}$) Coefficient of pg^2 in cost term.
cost_linear: ($c_{1k}$) Coefficient of pg in cost term.
cost_offset: ($c_{0k}$) Constant coefficient in cost term.

- grid.nodes.load:

pd: ($\Re(S_k^d)$) Real power demand (perturbed in dataset).
qd: ($\Im(S_k^d)$) Reactive power demand (perturbed in dataset).

- grid.nodes.shunt:

bs: ($\Im(Y_k^s)$) Shunt susceptance.
gs: ($\Re(Y_k^s)$) Shunt conductance.

- grid.edges.ac_line.features:

angmin: ($\theta_{ij}^{\Delta l}$) Minimum angle difference between from and to bus (radians).
angmax: ($\theta_{ij}^{\Delta u}$) Maximum angle difference between from and to bus (radians).
b_fr: ($\Im(Y_{ij}^c)$) Line charging susceptance at from bus.
b_to: ($\Im(Y_{ji}^c)$) Line charging susceptance at to bus.
br_r: ($\Re(1/Y_{ij})$) Branch series resistance.
br_x: ($\Im(1/Y_{ij})$) Branch series reactance.
rate_a: ($s_{ij}^u$) Long term thermal line rating.
rate_b: Short term thermal line rating.
rate_c: Emergency thermal line rating.

- grid.edges.transformer.features:

angmin: ($\theta_{ij}^{\Delta l}$) Minimum angle difference between from and to bus.
angmax: ($\theta_{ij}^{\Delta u}$) Maximum angle difference between from and to bus.
br_r: ($\Re(1/Y_{ij})$) Branch series resistance.
br_x: ($\Im(1/Y_{ij})$) Branch series reactance.
rate_a: ($s_{ij}^u$) Long term thermal line rating.
rate_b: Short term thermal line rating.
rate_c: Emergency thermal line rating.
tap: ($|T_{ij}|$) Branch off nominal turns ratio.
shift: ($\angle T_{ij}$) Branch phase shift angle.
b_fr: ($\Im(Y_{ij}^c)$) Line charging susceptance at from bus.
b_to: ($\Im(Y_{ji}^c)$) Line charging susceptance at to bus.

- grid.context:

baseMVA: The system wide MVA value for converting between mixed-units and p.u. unit values.

- solution.nodes.bus:

va: ($\angle(V_i V_j^*)$) Voltage angle.
vm: ($|V_i|$) Voltage magnitude.

- solution.nodes.generator:

pg: ($\Re(S_k^g)$) Real power generation.
qg: ($\Im(S_k^g)$) Reactive power generation.

- solution.edges.ac_line.features:

pt: ($\Re(S_{ji})$) Active power withdrawn at the to bus.
qt: ($\Im(S_{ji})$) Reactive power withdrawn at the to bus.
pf: ($\Re(S_{ij})$) Active power withdrawn at the from bus.
qf: ($\Re(S_{ij})$) Reactive power withdrawn at the from bus.

- `solution.edges.transformer.features`:
  - `pt`: Active power withdrawn at the to bus.
  - `qt`: Reactive power withdrawn at the to bus.
  - `pf`: Active power withdrawn at the from bus.
  - `qf`: Reactive power withdrawn at the from bus.
- `metadata`:
  - `objective`: AC-OPF objective achieved by conventional solver ($/h).

**OPFData**: Large-scale datasets for AC optimal power flow with topological perturbationsxsets:

- $\mathcal{N}$ - buses
- $\mathcal{R}$ - reference buses
- $\mathcal{E}, \mathcal{E}^R$ - branches, forward and reverse orientation
- $\mathcal{G}, \mathcal{G}_i$ - generators and generators at bus $i$
- $\mathcal{L}, \mathcal{L}_i$ - loads and loads at bus $i$
- $\mathcal{S}, \mathcal{S}_i$ - shunts and shunts at bus $i$

data:

- $S_k^{gl}, S_k^{gu} \ \forall k \in \mathcal{G}$ - generator complex power bounds
- $c_{2k}, c_{1k}, c_{0k} \ \forall k \in \mathcal{G}$ - generator cost components
- $v_i^l, v_i^u \ \forall i \in \mathcal{N}$ - voltage bounds
- $S_k^d \ \forall k \in \mathcal{L}$ - load complex power consumption
- $Y_k^s \ \forall k \in \mathcal{S}$ - bus shunt admittance
- $Y_{ij}, Y_{ij}^c, Y_{ji}^c \ \forall (i,j) \in \mathcal{E}$ - branch pi-section parameters
- $T_{ij} \ \forall (i,j) \in \mathcal{E}$ - branch complex transformation ratio
- $s_{ij}^u \ \forall (i,j) \in \mathcal{E}$ - branch apparent power limit
- $\theta_{ij}^{\Delta l}, \theta_{ij}^{\Delta u} \ \forall (i,j) \in \mathcal{E}$ - branch voltage angle difference bounds

variables:

- $S_k^g \ \forall k \in \mathcal{G}$ - generator complex power dispatch
- $V_i \ \forall i \in \mathcal{N}$ - bus complex voltage
- $S_{ij} \ \forall (i,j) \in \mathcal{E} \cup \mathcal{E}^R$ - branch complex power flow

minimize: $\sum_{k \in \mathcal{G}} c_{2k}(\Re(S_k^g))^2 + c_{1k}\Re(S_k^g) + c_{0k}$

subject to:

$$\angle V_r = 0 \ \ \forall r \in \mathcal{R}$$

$$S_k^{gl} \leq S_k^g \leq S_k^{gu} \ \ \forall k \in \mathcal{G}$$

$$v_i^l \leq |V_i| \leq v_i^u \ \ \forall i \in \mathcal{N}$$

$$\sum_{k \in \mathcal{G}_i} S_k^g - \sum_{k \in \mathcal{L}_i} S_k^d - \sum_{k \in \mathcal{S}_i} (Y_k^s)^* |V_i|^2 = \sum_{(i,j) \in \mathcal{E}_i \cup \mathcal{E}_i^R} S_{ij} \ \ \forall i \in \mathcal{N}$$

$$S_{ij} = \left(Y_{ij} + Y_{ij}^c\right)^* \frac{|V_i|^2}{|T_{ij}|^2} - Y_{ij}^* \frac{V_i V_j^*}{T_{ij}} \ \ \forall (i,j) \in \mathcal{E}$$

$$S_{ji} = \left(Y_{ij} + Y_{ji}^c\right)^* |V_j|^2 - Y_{ij}^* \frac{V_i^* V_j}{T_{ij}^*} \ \ \forall (i,j) \in \mathcal{E}$$

$$|S_{ij}| \leq s_{ij}^u \ \ \forall (i,j) \in \mathcal{E} \cup \mathcal{E}^R$$

$$\theta_{ij}^{\Delta l} \leq \angle(V_i V_j^*) \leq \theta_{ij}^{\Delta u} \ \ \forall (i,j) \in \mathcal{E}$$

Figure 1 | AC-OPF problem formulation, from PowerModels.jl [38]. Note: the branch current limit constraint is not included here as this is not present in the pglib cases.